\documentclass[a4paper,10pt]{article}
\usepackage[english,spanish]{babel}
\usepackage[utf8]{inputenc}
\usepackage{epigraph}
\usepackage{array}  
\usepackage{tikz}
\usetikzlibrary{babel}  
\usepackage{tikz-qtree}
\usepackage{tikz-dependency}
\usetikzlibrary{automata}
\usetikzlibrary{shapes.arrows,chains,positioning}
\usepackage[all]{xy}

\newcommand{\ra}{\rightarrow}

\title{El Lenguaje Natural como Lenguaje Formal}

\author{Franco M. Luque}
\date{}

\begin{document}

\maketitle

\begin{otherlanguage}{english}
\begin{abstract}
Formal languages theory is useful for the study of natural language.
In particular, it is of interest to study the adequacy of the grammatical
formalisms to express syntactic phenomena present in natural language.
First, it helps to draw hypothesis about the nature and complexity of the
speaker-hearer linguistic competence, a fundamental question in linguistics and
other cognitive sciences. Moreover, from an engineering point of view, it allows
the knowledge of practical limitations of applications based on those
formalisms.
In this article I introduce the adequacy problem of grammatical formalisms for
natural language, also introducing some formal language theory concepts required
for this discussion. Then, I review the formalisms that have been proposed in
history, and the arguments that have been given to support or reject their
adequacy.
\end{abstract}
\end{otherlanguage}

\begin{abstract}
La teor\'ia de lenguajes formales es \'util para el estudio de los lenguajes
naturales.
En particular, resulta de inter\'es estudiar la adecuaci\'on de los formalismos
gramaticales para expresar los fen\'omenos sint\'acticos presentes en el
lenguaje natural.
Primero, ayuda a trazar hip\'otesis acerca de la naturaleza y complejidad de
las competencias ling\"u\'isticas de los hablantes-oyentes del lenguaje, un
interrogante fundamental de la ling\"u\'istica y otras ciencias cognitivas.
Además, desde el punto de vista de la ingenier\'ia, permite conocer limitaciones
pr\'acticas de las aplicaciones basadas en dichos formalismos.
En este art\'iculo hago una introducci\'on al problema de la adecuaci\'on de los
formalismos gramaticales para el lenguaje natural, introduciendo tambi\'en
algunos conceptos de la teor\'ia de lenguajes formales necesarios para esta
discusi\'on.
Luego, hago un repaso de los formalismos que han sido propuestos a lo largo de
la historia, y de los argumentos que se han dado para sostener o refutar su
adecuaci\'on.
\end{abstract}

\newpage

\section{Introducción}

\epigraph{%
\textit{``One morning I shot an elephant in my pajamas.
How he got into my pajamas I don't know.''}%
}{%
Groucho Marx, \textit{Animal Crackers} (1930)%
}
\epigraph{%
\textit{``Our three weapons are fear, surprise, ruthless efficiency, and an almost
fanatical devotion to the Pope.''}%
}{%
Monty Python, \textit{The Spanish Inquisition} (1970)%
}

Noam Chomsky, en la década del 50, sentó las bases de la lingüística moderna al
empezar a estudiar la sintaxis del lenguaje utilizando herramientas matemáticas.
En sus primeros trabajos, dió inicio a la discusión acerca del lugar en el que
los lenguajes naturales se sitúan dentro de la denominada jerarquía de
lenguajes formales de Chomsky.
Esta discusión continuó luego por varias décadas, provocando incluso la proposición
de numerosos formalismos gramaticales nuevos.

En este artículo hago una introducción al problema de la adecuación de los
formalismos gramaticales para el lenguaje natural, introduciendo también algunos
conceptos elementales de la teoría de lenguajes formales, necesarios para esta
discusión.
Luego, hago un repaso de los diferentes formalismos gramaticales que
han sido propuestos a lo largo de la historia, y de los argumentos que se han
dado para sostener o refutar la adecuación de cada uno de éstos.

%


El artículo se encuentra estructurado como sigue.
En la siguiente sección se hace una breve introducción a la teoría de lenguajes
formales, a la notación a utilizar y a dos mecanismos gramaticales básicos para
la definición de lenguajes.
A continuación, en la sección~\ref{sec:methodology}, se define un marco metodológico
a la discusión.
En la sección~\ref{sec:natural}, se aborda la discusión de la adecuación de los
diferentes formalismos gramaticales, recorriendo la jerarquía de Chomsky en
orden creciente de expresividad.
La sección~\ref{sec:discussion} finaliza el artículo con algunas
reflexiones acerca de las implicancias para las áreas de la Lingüística
Computacional y el Procesamiento de Lenguaje Natural.

\section{Teoría de Lenguajes Formales}
\label{sec:formal}

En la teoría de lenguajes formales, un \textit{lenguaje} sobre un alfabeto
$\Sigma$ es un conjunto de secuencias, posiblemente infinito, a donde
cada secuencia se compone de una cantidad finita de símbolos tomados del
alfabeto $\Sigma$.

El conjunto de todas las posibles palabras que se pueden formar con un
alfabeto $\Sigma$ se denota $\Sigma^*$.
La secuencia vacía también es un elemento posible de un lenguaje y se denota
con la letra griega $\lambda \in \Sigma^*$.
Usualmente, usamos letras del comienzo del abecedario ($a, b, c, \ldots$) para
representar elementos de $\Sigma$, y del final del alfabeto ($r, s, t, u, v,
\ldots$) para representar elementos de $\Sigma^*$.
Escribimos $r^n$, con $n \geq 0$, para representar la secuencia que resulta de
repetir $n$ veces la secuencia $r$.
%
%

Un lenguaje puede ser definido por extensión, como por ejemplo el lenguaje
$L = \{ a, aba, bab \}$, sobre $\Sigma = \{a¸ b\}$, o por comprensión, como por
ejemplo $L = \{ a^n b^n | n \geq 0 \}$, que es el conjunto de todas las
secuencias que tienen primero cierta cantidad de letras $a$, seguidas de la
misma cantidad de letras $b$.
Por supuesto, un lenguaje infinito no puede ser definido por extensión.

%
Existen diversos mecanismos formales para la definición de lenguajes por
comprensión más allá de la notación clásica de conjuntos, que llamaremos
\textit{formalismos gramaticales}.
Cada formalismo gramatical tiene asociado una \textit{clase de lenguajes}, esto
es, el conjunto de todos lenguajes que pueden ser definidos usando tal formalismo.
A más \textit{poder expresivo}, más grande es la clase de lenguajes asociada.

\subsection{La Jerarquía de Chomsky}
\label{sec:hierarchy}

%
Chomsky estudió diversos formalismos gramaticales desde una prespectiva
lingüística.
En \cite{chomsky56}, por primera vez, describió tres modelos formales de
creciente expresividad, y estudió la adecuación de cada uno de ellos para
explicar la sintaxis del idioma inglés.
Más tarde, en \cite{chomsky59}, abordó un cuarto modelo formal aún más
expresivo, las Máquinas de Turing, para luego definir los tres modelos
anteriores como versiones cada vez más restringidas de éste.

\begin{figure}[tp]
 \centering
\begin{tikzpicture}
  \draw (0, 0) ellipse (1.5 and 0.5);
  \draw (0, 0.5) ellipse (2.5 and 1);
  \draw (0, 1) ellipse (3.5 and 1.5);
  \draw (0, 1.5) ellipse (4.5 and 2);
  \draw (0, 0) node {regulares};
  \draw (0, 1) node {libres de contexto};
  \draw (0, 2) node {sensibles al contexto};
  \draw (0, 3) node {recursivamente enumerables};
\end{tikzpicture}
 \caption{Relación de inclusión entre las cuatro clases de lenguajes de la
 jerarquía de Chomksy mostradas en el Cuadro~\ref{tab:hierarchy}.}
 \label{fig:hierarchy}
\end{figure}
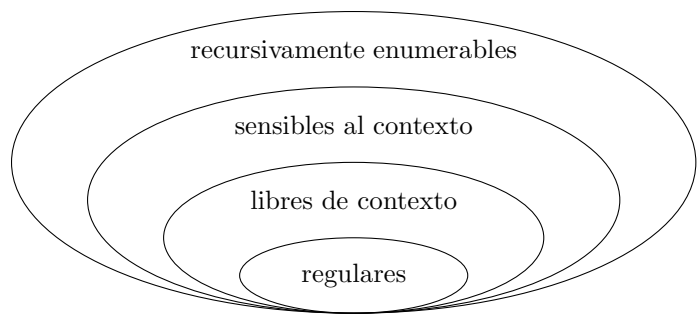

Los cuatro modelos propuestos por Chomsky componen una jerarquía de formalismos
llamada \textit{jerarquía de Chomsky}%
\footnote{A veces llamada jerarquía de Chomsky–Schützenberger, esta jerarquía
no puede ser atribuída únicamente a Chomsky, ya que muchos autores participaron
en el estudio y la proposición de los formalismos que la componen.}.
En el Cuadro~\ref{tab:hierarchy} se muestra la jerarquía, mientras que en la
Fig.~\ref{fig:hierarchy} se puede ver la relación de inclusión que existe entre
las clases de lenguajes correspondientes.

\begin{table}[tp]
 \centering
\begin{tabular}{ll}
Clase de Lenguajes & Formalismo Gramatical \\
\hline
Recursivamente Enumerables (REs) & Máquinas de Turing \\
Sensibles al Contexto (CSLs) & Gramáticas Sensibles al Contexto (CSGs) \\
Libres de Contexto (CFLs) & Gramáticas Libres de Contexto (CFGs) \\
Regulares & Autómatas Finitos Deterministas (DFAs)
 \end{tabular}
 \caption{Las cuatro clases de lenguajes de la jerarquía de Chomsky, y
 formalismos gramaticales repesentativos para cada una de ellas.}
 \label{tab:hierarchy}
\end{table}

En las siguientes secciones introducimos los dos formalismos más simples dentro
de la jerarquía: los autómatas finitos deterministas y las gramáticas libres de
contexto.

\subsection{Autómatas Finitos Deterministas}
\label{sec:dfas}


Un \textit{Autómata Finito Determinista} (DFA, por sus siglas en inglés) es una
máquina de estados que emite un símbolo del alfabeto cada vez que se realiza una
transición de un estado a otro.
Siempre comienza en un único estado denominado \textit{estado inicial}, y
termina en cualquier estado que pertenezca a un conjunto de \textit{estados
finales}.
Luego, un DFA se define por los siguientes elementos:
\begin{itemize}
 \item Un alfabeto finito $\Sigma$,
 \item un conjunto finito de estados $Q$,
 \item un estado inicial $q_0 \in Q$,
 \item un conjunto de estados finales $F \subseteq Q$ y
 \item una función de transición $\delta : Q \times \Sigma \rightarrow Q$, tal
 que $\delta(p, a) = q$ indica que la transición que parte del estado $p$ emitiendo
 el símbolo $a$ tiene como destino el estado $q$. Gráficamente:
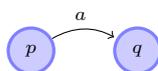
\begin{center}
   \begin{tikzpicture}[shorten >=1pt,node distance=1.4cm,on grid,auto,font=\scriptsize]
    \tikzstyle{every state}=[draw=blue!50,very thick,fill=blue!20,inner
sep=0pt,minimum size=6mm]
    \node[state] (p) {$p$};
    \node[state] (q) [right of=p] {$q$};
    \path[->] 
      (p) edge [bend left] node [align=center] {$a$} (q);
\end{tikzpicture}
\end{center}
 \end{itemize}

El lenguaje definido por un DFA es el conjunto de sequencias que pueden ser
generadas en cualquier recorrido del autómata que empiece en el estado inicial y
termine en un estado final.
Un DFA puede definir un lenguaje infinito si contiene ciclos en sus
transiciones.

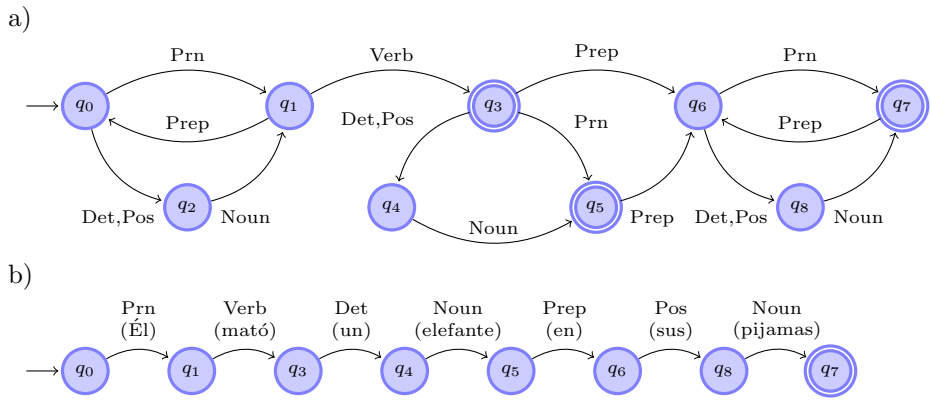
\begin{figure}[tp]
 \centering
\begin{tabular}{@{}l@{}c@{}}
a) \\
  \begin{tikzpicture}[shorten >=1pt,node distance=1.9cm,on grid,auto,font=\scriptsize]
    \tikzstyle{every state}=[draw=blue!50,very thick,fill=blue!20,inner
sep=0pt,minimum size=6mm]
    \node[state,initial,initial text=] (q_0) {$q_0$};
    \node[state] (q_2) [below right of=q_0] {$q_2$};
    \node[state] (q_1) [above right of=q_2] {$q_1$};
    \node[state] (q_4) [below right of=q_1] {$q_4$};
    \node[state,accepting] (q_3) [above right of=q_4] {$q_3$};
    \node[state,accepting] (q_5) [below right of=q_3] {$q_5$};
    \node[state] (q_6) [above right of=q_5] {$q_6$};
    \node[state] (q_8) [below right of=q_6] {$q_8$};
    \node[state,accepting] (q_7) [above right of=q_8] {$q_7$};
    \path[->] 
      (q_0) edge [bend left] node {Prn} (q_1)
      (q_0) edge [bend right] node [swap,at end] {Det,Pos} (q_2)
      (q_1) edge [bend left] node [swap] {Prep} (q_0)
      (q_2) edge [bend right] node [swap,at start] {Noun} (q_1)
      (q_1) edge [bend left] node {Verb} (q_3)
      (q_3) edge [bend right] node [swap] {Det,Pos} (q_4)
      (q_3) edge [bend left] node {Prn} (q_5)
      (q_3) edge [bend left] node {Prep} (q_6)
      (q_4) edge [bend right] node {Noun} (q_5)
      (q_5) edge [bend right] node [swap,at start] {Prep} (q_6)
      (q_6) edge [bend left] node {Prn} (q_7)
      (q_6) edge [bend right] node [swap,at end] {Det,Pos} (q_8)
      (q_7) edge [bend left] node [swap] {Prep} (q_6)
      (q_8) edge [bend right] node [swap,at start] {Noun} (q_7)
      ;
  \end{tikzpicture}
\\ b) \\
  \begin{tikzpicture}[shorten >=1pt,node distance=1.4cm,on grid,auto,font=\scriptsize]
    \tikzstyle{every state}=[draw=blue!50,very thick,fill=blue!20,inner
sep=0pt,minimum size=6mm]
    \node[state,initial,initial text=] (q_0) {$q_0$};
    \node[state] (q_1) [right of=q_0] {$q_1$};
    \node[state] (q_3) [right of=q_1] {$q_3$};
    \node[state] (q_4) [right of=q_3] {$q_4$};
    \node[state] (q_5) [right of=q_4] {$q_5$};
    \node[state] (q_6) [right of=q_5] {$q_6$};
    \node[state] (q_8) [right of=q_6] {$q_8$};
    \node[state,accepting] (q_7) [right of=q_8] {$q_7$};
    \path[->] 
      (q_0) edge [bend left] node [align=center] {Prn\\(Él)} (q_1)
      (q_1) edge [bend left] node [align=center] {Verb\\(mató)} (q_3)
      (q_3) edge [bend left] node [align=center] {Det\\(un)} (q_4)
      (q_4) edge [bend left] node [align=center] {Noun\\(elefante)} (q_5)
      (q_5) edge [bend left] node [align=center] {Prep\\(en)} (q_6)
      (q_6) edge [bend left] node [align=center] {Pos\\(sus)} (q_8)
      (q_8) edge [bend left] node [align=center] {Noun\\(pijamas)} (q_7)
      ;
  \end{tikzpicture}
\end{tabular}
\caption{
a)
Ejemplo de autómata finito determinista (DFA) para un lenguaje natural de
juguete.
Por simplicidad, se describe un lenguaje de categorías léxicas en lugar de un
lenguaje léxico.
No se dibujan aquellas transiciones que no pueden conducir a un estado final.
b)
Recorrido de aceptación de la secuencia de ejemplo
``Prn Verb Det Noun Prep Pos Noun''.
}
\label{fig:toydfa}
\end{figure}

Un ejemplo gráfico de un DFA se muestra en la Fig.~\ref{fig:toydfa}a.
El alfabeto de este autómata es $\Sigma = \{$Prn, Det, Noun, Verb, Prep, Pos$\}$.
El estado inicial es $q_0$, como se indica con la flecha que lo apunta, y los
estados finales son aquellos que tienen doble contorno.
Por simplicidad, no se dibujan las transiciones que no pueden conducir a un
estado final.%
\footnote{Estas transiciones en realidad tienen como destino un estado
adicional, tampoco dibujado, que es no-final y cuyas transiciones vuelven al
mismo estado.}
En la Fig.~\ref{fig:toydfa}b puede verse el recorrido que muestra que la
secuencia ``Prn Verb Det Noun Prep Pos Noun'' es generada por el DFA.


El conjunto de todos los lenguajes posibles que pueden ser generados por DFAs
constituye la clase de \textit{lenguajes regulares}.
Existen otros formalismos con el mismo poder expresivo, como los autómatas
finitos no-deterministas o las expresiones regulares.

Hay, por supuesto, muchos lenguajes que no son regulares, es decir, que se sabe
que no existe ningún DFA que los generen.
Un ejemplo arquetípico de lenguaje no regular es el ya mencionado
$L = \{ a^n b^n | n \geq 0 \}$.

Existe al menos una manera de demostrar que un lenguaje no es regular, que es
usando el denominado \textit{pumping lemma}.
Este lema hace uso del hecho de que si un lenguaje es regular y al mismo tiempo
infinito, el DFA que lo genera obligatoriamente debe tener un ciclo en un camino
a un estado final.
Haciendo uso de este ciclo se pueden generar infinitas secuencias que
obligatoriamente deben pertenecer al lenguaje.

Por ejemplo, se puede probar que si $L = \{ a^n b^n | n \geq 0 \}$ es regular,
usando el \textit{pumping lemma} necesariamente debe ser posible generar una
secuencia $a^n b^m$ con $n \neq m$, y por lo tanto esta secuencia debe
pertenecer al lenguaje $L$.
Como esto es absurdo, $L$ no es regular.

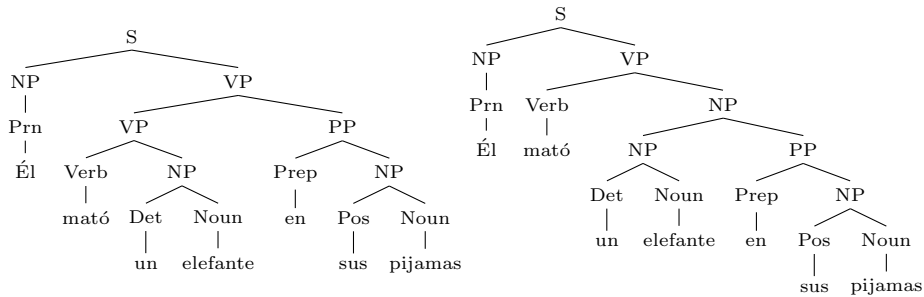
\begin{figure}[tp]
 \centering
\begin{tabular}{@{}l@{}c@{}}
a) \\ &
{
\small
\begin{tabular}{ll}
S $\ra$ NP VP & (oraciones) \\
NP $\ra$ Prn $\mid$ Det Noun $\mid$ Pos Noun $\mid$ NP PP & (sintagmas nominales) \\
VP $\ra$ Verb NP $\mid$ VP PP & (sintagmas verbales) \\
PP $\ra$ Prep NP & (sintagmas preposicionales) \\
Prn $\ra$ Él $\mid$ $\ldots$ & (pronombres) \\
Det $\ra$ un $\mid$ $\ldots$ & (determinantes) \\
Noun $\ra$ elefante $\mid$ pijamas $\mid$ $\ldots$ & (sustantivos) \\
Verb $\ra$ mató $\mid$ $\ldots$ & (verbos) \\
Prep $\ra$ en $\mid$ $\ldots$ & (preposiciones) \\
Pos $\ra$ sus $\mid$ $\ldots$ & (posesivos)
\end{tabular}
}
%
\\ b) \\ &
\begin{tabular}{@{}m{0.5\textwidth}@{}m{0.5\textwidth}@{}}
{
\scriptsize
\begin{tikzpicture}
\tikzset{level distance=0.6cm}
      \Tree [.S
	      [.NP [.Prn Él ] ]
	      [.VP
		  [.VP
		    [.Verb mató ]
		      [.NP
			[.Det un ]
			[.Noun elefante ]
		      ]
		  ]
		  [.PP
		    [.Prep en ]
		    [.NP
		      [.Pos sus ]
		      [.Noun pijamas ]
		    ]
		  ]
	      ] ]
\end{tikzpicture}
}
&
{
\scriptsize
\begin{tikzpicture}
\tikzset{level distance=0.6cm}
      \Tree [.S
	      [.NP [.Prn Él ] ]
	      [.VP
		  [.Verb mató ]
		  [.NP
		    [.NP
		      [.Det un ]
		      [.Noun elefante ]
		    ]
		    [.PP
		      [.Prep en ]
		      [.NP
			[.Pos sus ]
			[.Noun pijamas ]
		      ]
		    ]
		  ]
	      ] ]
\end{tikzpicture}
}
\end{tabular}
\end{tabular}
\caption{
a) Ejemplo de gramática libre de contexto (CFG) para un lenguaje natural de juguete.
b) Dos análisis posibles para la oración ``Él mató un elefante en sus
pijamas'' de acuerdo a esta CFG.
}
\label{fig:toycfg}
\end{figure}


\subsection{Gramáticas Libres de Contexto}
\label{sec:cfgs}

Una \textit{Gramática Libre de Contexto} (CFG, por sus siglas en inglés) es un
sistema que permite construir secuencias a partir de la aplicación repetida de
reglas.
Las reglas permiten reemplazar símbolos no-terminales por secuencias de
terminales (símbolos del alfabeto) y no-terminales.
%
Luego, una CFG se define por los siguientes elementos:
\begin{itemize}
 \item un alfabeto finito $\Sigma$ (terminales),
 \item un conjunto finito de no-terminales $N$ (también llamados estados),
 \item un no-terminal inicial $S \in N$ y
 \item un conjunto de reglas $R$, que escribimos de la forma $X \ra \alpha$, a
 donde $X \in N$ y $\alpha \in (\Sigma \cup N)^*$ es una secuencia de terminales
 y no-terminales.
\end{itemize}

El lenguaje definido por una CFG es el conjunto de secuencias de terminales que
resulta de cualquier derivación que comienze desde el no-terminal inicial $S$.
Puede suceder que una misma secuencia pueda ser derivada de varias maneras.
En este caso, se dice que la CFG es ambigua.

Un ejemplo de CFG sobre el alfabeto $\Sigma = \{ a, b\}$ es el que tiene un
único no-terminal $S$ y el conjunto de reglas
$$R = \{ S \rightarrow \lambda, S \rightarrow a \; S \; b \}.$$
El lenguaje generado por esta CFG es $L = \{ a^n b^n | n \geq 0 \}$.
Puede verse que la segunda regla es recursiva y por lo
tanto puede aplicarse cíclicamente.

Otro ejemplo de CFG se puede ver en la Fig.~\ref{fig:toycfg}a.
Por brevedad, las reglas se muestran agrupadas por no-terminal del lado
izquierdo, uniendo los diferentes lados derechos con el conector $\mid$.
En las Fig.~\ref{fig:toycfg}b se pueden ver dos derivaciones diferentes para
una misma secuencia, por lo que esta CFG es ambigua.

% ...
El conjunto de todos los lenguajes que pueden ser generados por CFGs se denomina
la clase de \textit{lenguajes libres de contexto} (CFLs).
Esta clase incluye a todos los lenguajes regulares y también contiene lenguajes
no-regulares, es decir, las CFGs tienen un mayor poder expresivo que los DFAs.

Así como para los lenguajes regulares, también para los lenguajes libres de
contexto existe una versión del \textit{pumping lemma} que se puede usar para
demostrar que un lenguaje no es libre de contexto.
Por ejemplo, se puede demostrar por el absurdo que
$L = \{ a^n b^n c^n | n \geq 0 \}$ no es libre de contexto ya que, si lo fuera,
por el \textit{pumping lemma} también deberían pertenecer al lenguaje secuencias
en los que la cantidad de $a$'s, $b$'s y $c$'s no coinciden.

\section{Consideraciones Metodológicas}
\label{sec:methodology}

El estudio del lenguaje natural como un lenguaje formal requiere de un conjunto
de suposiciones y definiciones previas.
La mayor parte de ellas se remontan a los orígenes de la lingüística generativa
de Chomsky
(\cite{chomsky57}, pp. 13-17, \cite{chomsky65}, pp. 3-4), y se sostienen hasta
el día de hoy en las principales ramas de la lingüística.
En esta sección repasamos brevemente las suposiciones y definiciones más
relevantes.

\subsection{Gramaticalidad vs. Aceptabilidad}
\label{sec:grammaticality}

El primer conjunto de suposiciones se refiere a una idealización de las
capacidades lingüísticas humanas.
A la hora de discutir el lenguaje natural, se asume un \textit{hablante-oyente
ideal} sin limitaciones de memoria, distracciones o errores.
Además, se asume que los lenguajes naturales tienen un comportamiento estático
en el tiempo y el espacio:
No cambian ni evolucionan, ni varían dentro de la comunidad de hablantes.

En este marco, la \textit{gramática generativa} de un lenguaje natural es la
descripción de la competencia lingüística del hablante-oyente ideal de ese
lenguaje.
Así, podemos hablar de la \textit{gramaticalidad} de una oración para indicar
que ésta puede ser explicada por las reglas de la gramática.
En este sentido, suponemos que se puede identificar a un lenguaje natural con el
conjunto de oraciones gramaticales que la componen.

Siendo la gramática un objeto ideal e inasequible, sólo se puede recurrir a
métodos indirectos para determinar la gramaticalidad de las oraciones.
Uno de los principales recursos metodológicos que se ha utilizado en la
literatura es el juicio intuitivo de hablantes-oyentes competentes.

El concepto de gramaticalidad puede verse en contraste con el de
\textit{aceptabilidad}, que se refiere a cuán aceptable es una oración para
un hablante-oyente no ideal al margen del estricto cumplimiento de reglas
gramaticales.
En la aceptabilidad intervienen otros factores como simplicidad,
comprensibilidad, frecuencia, etc.

Un ejemplo clásico que permite ilustrar la diferencia entre gramaticalidad y
aceptabilidad es la famosa oración de Chomsky \cite{chomsky57}
\begin{quote}
\textit{Colorless green ideas sleep furiously.}

(Ideas verdes incoloras duermen furiosamente.)
\end{quote}
Esta oración es inaceptable por carecer de significado pero se corresponde con
las reglas gramaticales del idioma inglés y es consistentemente juzgada como
gramatical por los hablantes-oyentes del idioma.

\subsection{Adecuación Débil vs. Adecuación Fuerte}
\label{sec:adequacy}

Existe una distinción importante a la hora de considerar la adecuación de
formalismos gramaticales para expresar el lenguaje natural (\cite{chomsky65}, pp. 60-62).
%
La \textit{adecuación débil} es la capacidad de un formalismo de
expresar el conjunto de oraciones gramaticalmente correctas de los lenguajes
naturales.
Esta adecuación no garantiza que el formalismo sea capaz de dar las
descripciones estructurales correctas de las oraciones.
La \textit{adecuación fuerte} se refiere a esta última capacidad.

La relevancia lingüística de la distinción entre adecuación débil y adecuación
fuerte puede ser apreciada con el siguiente ejemplo.
Consideremos la gramática de la Fig.~\ref{fig:toycfg}, que corresponde a una
versión simplificada del idioma inglés.
A pesar de que se trata de una CFG, el lenguaje generado es regular ya que
también puede ser generado por el DFA de la Fig.~\ref{fig:toydfa}a.%
\footnote{Por supuesto, la afirmación de que la CFG y el DFA generan el mismo
lenguaje requiere de una demostración matemática.
}
Luego, tanto los DFAs como las CFGs son débilmente adecuadas para expresar este
idioma de juguete.
Por otro lado, en términos de adecuación fuerte, la representación que ofrece el
DFA no es admisible, ya que no permite expresar algunas propiedades
estructurales del inglés.
Por ejemplo, para la oración 
\begin{quote}
Él mató a un elefante en sus pijamas.
\end{quote}
la adecuación fuerte requiere una manera de expresar la ambigüedad de que la
frase preposicional ``en sus pijamas'' puede afectar o bien al verbo (matar) o
bien al objeto (el elefante).%
\footnote{
Este ejemplo se deriva de la frase de Groucho Marx citada al comienzo
del artículo.
La gracia de la frase reside justamente en la ambigüedad explicada.
}
Las CFGs son capaces de hacer esta distinción permitiendo la asignación de
estructuras ambiguas para la oración, como se muestra en la
Fig.~\ref{fig:toycfg}b.

\subsection{Tratabilidad y Complejidad}

Un criterio adicional surge ya desde Chomsky \cite{chomsky65}, al
incluir como requisito para la adecuación la existencia de un método
para resolver el problema de obtener la descripción estructural de una oración dada.
Nos referimos a este problema como el \textit{problema de análisis
(parsing)}.
Un problema asociado al problema de parsing es el \textit{problema de
reconocimiento}, que se trata de decidir si una oración dada es gramatical.

En el ámbito de los lenguajes formales, denominamos \textit{tratabilidad} a la
existencia de soluciones algorítmicas para resolver estos problemas.
Además de la \textit{tratabilidad}, también nos interesa la \textit{complejidad}
de los algoritmos, esto es, la cantidad de tiempo y espacio de cómputo que
requieren para solucionar los problemas.


Para considerar adecuado un formalismo gramatical en cuanto a complejidad,
deben existir algoritmos eficientes para la solución de estos problemas.
En general, se considera aceptable una complejidad de \textit{orden polinomial},
esto es, que el tiempo y el espacio requerido por los algoritmos sean funciones
polinómicas en términos de los tamaños de la gramática y de la oración de entrada.




\section{El Lenguaje Natural como...}
\label{sec:natural}

La jerarquía de Chomsky es un buen punto de comienzo para determinar a dónde
entra el lenguaje natural dentro del mundo de los lenguajes formales.
En esta sección hacemos una recorrida de la jerarquía, revisando los argumentos
que se han dado en la literatura en cada caso para apoyar o refutar la
adecuación de los diferentes formalismos gramaticales.

\subsection{... Lenguaje Regular}

La primera pregunta a responder, y aparentemente la más fácil, es si los
lenguajes naturales son regulares.
En términos de adecuación fuerte, ya vimos en la sección~\ref{sec:adequacy} que
al menos los DFAs no permiten ofrecer descripciones estructurales ambiguas como
las que se presentan en el fenómeno de la adjunción de los sintagmas
preposicionales.
Sin embargo, esto no significa que no puedan existir otros mecanismos regulares
que sean capaces de hacer esto, como por ejemplo pueden los autómatas
no-deterministas.
De cualquier manera, la posiblidad de la adecuación fuerte de cualquier
mecanismo regular queda descartada al comprobar que los lenguajes regulares no
son ni siquiera débilmente adecuados.

Existen en la literatura muchas maneras de probar la no-regularidad del lenguaje
natural.
Chomsky (\cite{chomsky57}, pp. 21-22) presentó pruebas basadas en partes del
idioma inglés que toman la forma de lenguajes no-regulares como
$\{ a^n b^n | n \geq 0 \}$ o el lenguaje de las palabras capicúa
$\{ x | x \in \{a, b\}^*, x = reverse(x) \}$.


Partee (\cite{partee90}, pp. 480-482) elaboró una prueba basada en el fenónemo del
\textit{center-embedding} (subordinación central), que permite introducir
oraciones subordinadas en el medio de otras oraciones.
En el castellano, por ejemplo, el \textit{center-embedding} permite construir
las siguientes oraciones gramaticales:%
\footnote{Ejemplo basado en la canción infantil tradicional
``Sal de ahí, chivita, chivita''.}
\begin{quote}
La chiva murió.

La chiva, que el lobo sacó, murió.

La chiva, que el lobo, que el palo golpeó, sacó, murió.

La chiva, que el lobo, que el palo, que el fuego, quemó, golpeó, sacó, murió.%
\footnote{
Esta triple subordinación y subordinaciones de orden mayor, son ejemplos claros
de oraciones gramaticales, de acuerdo a la metodología adoptada, pero juzgadas
como no aceptables.}
\end{quote}
En general, podemos decir que son gramaticales todas las oraciones de la forma 
\begin{quote}
La chiva (, que $X$)$^n$ $(Y ,)^n$ murió.
\end{quote}
Aquí, $X \in \{$el lobo, el palo, el fuego, $\ldots\}$ son sintagmas nominales
(NPs), e $Y \in \{$sacó, golpeó, quemó, $\ldots\}$ son verbos transitivos.

Si el lenguaje que incluye todas estas oraciones fuera regular, por el
\textit{pumping lemma} para lenguajes regulares sería posible también construir
oraciones a donde la cantidad de NPs no coincide con la cantidad de verbos.
Es decir, el lenguaje obligatoriamente debería contener algunas oraciones de la
forma
\begin{quote}
La chiva (, que $X$)$^p$ $(Y ,)^q$ murió.
\end{quote}
con $p \neq q$.
Sin embargo, sabemos que todas estas oraciones son no-gramaticales.
Por ejemplo, son no gramaticales%
\footnote{En la literatura se acostumbra a marcar con * a los ejemplos de
oraciones no gramaticales.}
\begin{quote}
* La chiva sacó, murió. ($p = 0, q = 1$)

* La chiva, que el lobo murió. ($p = 1, q = 0$)
\end{quote}
Luego, este lenguaje no puede ser regular, ya que de serlo debería incluir
oraciones que no incluye.



\subsection{... Lenguaje Libre de Contexto}

La siguiente pregunta es acerca de la adecuación de los lenguajes libres de
contexto.
En términos de adecuación fuerte, Chomsky afirmó que las gramáticas libres de
contexto pueden expresar sólo torpemente algunas estructuras sintácticas simples
del inglés, como por ejemplo las conjunciones, los verbos auxiliares y la voz
pasiva (\cite{chomsky57}, pp. 34--43).

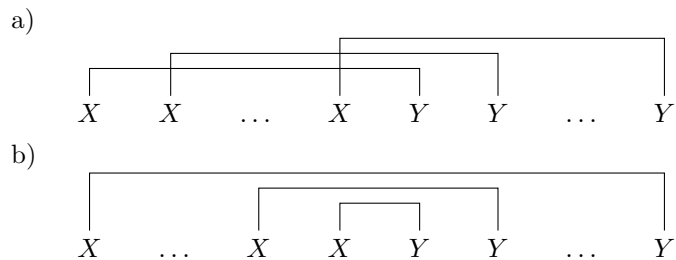
\begin{figure}[tp]
 \centering
\begin{tabular}{lc}
a) \\ &
 \begin{tikzpicture}[text height=1.5ex,
      node distance=.1cm and .5cm,
      nonterminal/.style={},
      skip loop/.style={to path={-- ++(0,#1) -| (\tikztotarget)}},
      point/.style={circle,inner sep=0pt,fill=red}
      ]
  \node (0) [nonterminal] {$X$};
  \node (1) [nonterminal,right= of 0] {$X$};
  \node (2) [nonterminal,right= of 1] {$\ldots$};
  \node (3) [nonterminal,right= of 2] {$X$};
  \node (4) [nonterminal,right= of 3] {$Y$};
  \node (5) [nonterminal,right= of 4] {$Y$};
  \node (6) [nonterminal,right= of 5] {$\ldots$};
  \node (7) [nonterminal,right= of 6] {$Y$};
  \path 
    (0) edge [-,skip loop=6mm] (4)
    (1) edge [-,skip loop=8mm] (5)
    (3) edge [-,skip loop=10mm] (7)
  ;
  \end{tikzpicture}
\\ b)  \\ &
 \begin{tikzpicture}[text height=1.5ex,
      node distance=.1cm and .5cm,
      nonterminal/.style={},
      skip loop/.style={to path={-- ++(0,#1) -| (\tikztotarget)}},
      point/.style={circle,inner sep=0pt,fill=red}
      ]
  \node (0) [nonterminal] {$X$};
  \node (1) [nonterminal,right= of 0] {$\ldots$};
  \node (2) [nonterminal,right= of 1] {$X$};
  \node (3) [nonterminal,right= of 2] {$X$};
  \node (4) [nonterminal,right= of 3] {$Y$};
  \node (5) [nonterminal,right= of 4] {$Y$};
  \node (6) [nonterminal,right= of 5] {$\ldots$};
  \node (7) [nonterminal,right= of 6] {$Y$};
  \path 
    (0) edge [-,skip loop=10mm] (7)
    (2) edge [-,skip loop=8mm] (5)
    (3) edge [-,skip loop=6mm] (4)
  ;
  \end{tikzpicture}
\end{tabular}
 \caption{
a) Esquema general de la estructura sintáctica de las interdependencias seriales
no acotadas presentes en los idiomas holandés y suizo-alemán.
Se puede ver cómo los elementos se asocian de manera intercalada.
b) Esquema general de la estructura sintáctica forzada por las gramáticas libre
de contexto.
Los elementos obligatoriamente deben asociarse desde el centro hacia afuera.
}
 \label{fig:dutch}
\end{figure}

Un argumento más conclusivo se dió muy posteriormente, entre los 70 y los 80,
en torno al fenómeno de \textit{interdependencias seriales no acotadas}
presente, por ejemplo, en los idiomas holandés y suizo-alemán \cite{Bresnan82}.
Este tipo de construcciones tienen la forma general $X^n Y^n$, parecida a la
subordinación central discutida en la sección anterior, pero a diferencia de
ésta, la estructura sintáctica asocia los elementos de manera intercalada,
como se muestra en la Fig.~\ref{fig:dutch}~a).
%
%
%
%
Las gramáticas libres de contexto pueden expresar lenguajes de la forma
$X^n Y^n$, pero no pueden hacerlo utilizando una estructura intercalada como la
que require este fenómeno sintáctico.
Sólo pueden hacerlo utilizando reglas con recursión central, forzando una
asociación de los elementos desde el centro hacia afuera, como se
muestra en la Fig.~\ref{fig:dutch}~b).


La discusión acerca de la adecuación débil de las gramáticas libres de contexto
ha sido mucho más complicada. 
Luego de que Chomsky planteara la pregunta en 1956, existieron muchos intentos
de demostrar la inadecuación débil, varios de ellos refutados por Pullum y
Gazdar en \cite{pullum82}.
Las fallas expuestas de las pruebas van desde errores matemáticos, como un uso
incorrecto del \textit{pumping lemma}, pasando por desacuerdos lingüísticos,
como la confusión entre fenómenos sintácticos y semánticos, hasta discusiones
metodológicas, como la manera en la que se decide la gramaticalidad de las
oraciones.

Un ejemplo de  argumento fallido es el que elaboró Elster \cite{elster78}
en torno a oraciones de la forma
\begin{quote}
Los primeros dos millones (de millones)$^p$ decimales de $\pi$ son $Y^q$
\end{quote}
a donde $Y$ es un dígito.
Según Elster, estas oraciones son gramaticales si y sólo si $p$ y $q$ son tales
que la cantidad de decimales en el predicado se corresponde con la indicada en
el sujeto ($q = 2 \times 10^{6(p+1)}$).
Si este lenguaje fuera libre de contexto, el \textit{pumping
lemma} para CFLs permitiría construir oraciones en las que $p$ y $q$ no se
corresponden, es decir, oraciones no gramaticales.
Luego, concluye Elster que el lenguaje natural no es libre de contexto.

Pullum y Gazdar \cite{pullum82} rechazaron de plano este argumento, sosteniendo
que la correspondencia entre $p$ y $q$ nada tienen que ver con la
gramaticalidad, así como tampoco presentan problemas de gramaticalidad las
siguientes oraciones:
\begin{quote}
A continuación hay seis números al azar: 3, 17, 8, 9, 41.

Nuestras tres armas son miedo, sorpresa, eficiencia despiadada, y una
devoción casi fanática por el Papa.%
\footnote{Traducción al castellano de la frase de los Monty Python citada al
comienzo del artículo.}
\end{quote}

No fue hasta 1985 que se propuso un argumento fuerte en contra de la adecuación
débil, dado por Shieber en \cite{Shieber85}.
Shieber construyó un contraejemplo tomado del dialecto suizo-alemán (o alemán de
Suiza), que combina interdependencias seriales, como las mencionadas
anteriormente para el idioma holandés, con el marcado de casos
(\textit{case-marking}) en los objetos, también presente en el idioma alemán.
En el \textit{case-marking} del suizo-alemán, los objetos se pueden marcar en
caso acusativo o dativo, y los verbos se dividen en subcategorías dependiendo
del caso que requieran para su objeto.

Por ejemplo, en la Fig.~\ref{fig:swiss-german} observamos una triple
interdependencia con un correcto marcado de casos.
Cada verbo en este ejemplo está asociado con un objeto del caso que le
corresponde.
En cambio, la oración
\begin{quote}
* ... mer d'chind \textbf{de} Hans es huus lönd hälfe aastriiche
\end{quote}
no es gramatical, ya que el sintagma ``de Hans'' está en caso acusativo pero el
verbo ``hälfe'' requiere un objeto de caso dativo (``em Hans'').

\begin{figure}[tp]
\small
 \centering
  \begin{dependency}[edge style={-},hide label]
  \tikzstyle{POS}=[font=\footnotesize]
  \begin{deptext}[column sep=0.2cm, row sep=0.5ex]
... mer \& d'chind \& em Hans \& es huus \& lönd \& hälfe \& aastriiche \\
... (nosotros) \& (los chicos) \& (a Hans) \& (la casa) \& (dejamos) \& (ayudar) \& (pintar) \\
\& |[POS]| ACC \& |[POS]| DAT \& |[POS]| ACC \\
  \end{deptext}
  \depedge[edge unit distance=.5ex]{2}{5}{}
  \depedge[edge unit distance=1ex]{3}{6}{}
  \depedge[edge unit distance=1.5ex]{4}{7}{}
  \end{dependency}
 \caption{Ejemplo en idioma suizo-alemán de interdependencias seriales y marcado
 de casos, para el fragmento de oración
 ``... nosotros dejamos a los chicos ayudar a Hans a pintar la casa''.
 ACC indica caso acusativo (artículos ``de'' y ``es'') y DAT caso dativo
 (artículo ``em'').}
 \label{fig:swiss-german}
\end{figure}


En general, una oración con interdependencias seriales es gramatical sí y sólo
si cada sintagma nominal está marcado con el caso que requiere el verbo que le
corresponde.
En particular, podemos agrupar todos los casos acusativos primero y luego todos
los casos dativos, para obtener oraciones con la forma general
\begin{quote}
$\ldots$ mer $X^n$ $Y^m$ es huus $Z^n$ $W^m$ asstriiche
\end{quote}
a dónde $X$ son NPs de caso acusativo (``d'chind''), $Y$ son NPs de caso dativo
(``em Hans''), $Z$ son verbos acusativos (``laa''), y $W$ son verbos dativos
(``hälfe'').
Si este lenguaje natural fuera libre de contexto, podría aplicarse el
\textit{pumping lemma} para CFLs para construir a partir de oraciones
gramaticales de este tipo, otras oraciones en las que la cantidad de objetos de
cada caso no se corresponde con la cantidad de verbos de cada subcategoría, es
decir, oraciones no gramaticales.
De esto se desprende la conclusión que este lenguaje natural no es libre de
contexto.


\subsection{... Lenguaje Sensible al Contexto, y Más Allá}

Los argumentos de inadecuación de los lenguajes libres de contexto exigieron
buscar un formalismo gramatical más expresivo para el lenguaje natural.
Yendo al siguiente escalón en la jerarquía de Chomsky, encontramos los lenguajes
sensibles al contexto (CSLs).


Los lenguajes sensibles al contexto son capaces de describir los fenómenos
sintácticos utilizados para probar la inadecuación libre de contexto, incluyendo
las interdependencias seriales del holandés, y las del suizo-alemán en
combinación con el marcado de casos.
En principio, no existen argumentos en la literatura que hablen en contra de
la adecuación débil o fuerte de los CSLs.

Sin embargo, los CSLs son inadecuados desde el punto de vista de
complejidad, una noción adicional de adecuación,
ya que se sabe que el problema de reconocimiento es PSPACE-complete para las
gramáticas sensibles al contexto \cite{hopcroft00}.
Esto significa que no se conocen, y posiblemente no existen%
\footnote{A no ser que P = PSPACE sea verdadero, uno de los problemas abiertos
más importantes de las ciencias de la computación.
La creencia mayoritaria de la comunidad científica es que P $\neq$ PSPACE.
}%
, algoritmos eficientes para determinar si una oración es gramatical o no, una
tarea que los humanos pueden hacer rápidamente con el lenguaje natural.


Formalismos aún más expresivos han sido propuestos como adecuados para el
lenguaje natural, como por ejemplo las gramáticas de unificación
(\textit{unification grammars}) \cite{francez11}, que pueden expresar la clase
entera de lenguajes recursivamente enumerables, al tope de la jerarquía de
Chomsky.
La altísima expresividad de estos formalismos indica que las adecuaciones débil
y fuerte no son un problema, pero sí se presenta un grave problema de
tratabilidad.
Los problemas de reconocimiento y análisis son en este caso indecidibles,
esto es, se sabe que no existen algoritmos que permitan resolverlos.

\subsection{... Lenguaje Moderadamente Sensible al Contexto}

El problema de la excesiva expresividad de los CSLs llevó a la búsqueda de
una nueva clase de lenguajes que se ubique en un paso intermedio entre los
libres de contexto y los sensibles al contexto.
La dificultad histórica que hubo para encontrar ejemplos de fenómenos
sintácticos genuinamente sensibles al contexto en el lenguaje natural, hizo
evidente que la expresividad requerida estaba apenas por encima de los CSLs.

Es por eso que en la literatura se propusieron nuevos formalismos gramaticales
que pudieran ser ligeramente más expresivos que las CFGs.
En \cite{joshi85}, Joshi introdujo las \textit{Tree Adjoining Grammars (TAGs)},
y mostró que éstas son capaces de expresar fenómenos como las interdependencias
seriales, y que lo hacen dando descripciones estructurales adecuadas.
%
Al mismo tiempo,
las TAGs tienen buenas propiedades de complejidad, parecidas a las de las CFGs,
ya que existen algoritmos polinomiales para los problemas de reconocimiento y
análisis.

Joshi definió vagamente el concepto de \textit{gramáticas moderadamente sensibles
al contexto (MCSGs)} como aquellas gramáticas que tienen esta capacidad
simultánea de describir interdependencias seriales y de tener \textit{parsing}
polinomial.%
\footnote{Además de la propiedad de crecimiento constante (\textit{constant
growth}), sobre la que hay cierta polémica y que por cuestiones de espacio
prefiero dejar fuera de la discusión. Ver \cite{radzinski91}, \cite{groenink97}
y \cite{kallmeyer10} (p. 2).}
Los TAGs, entonces, son una instancia particular de MCSG.
Varios mecanismos formales de aparición en la década de los 80 fueron
encontrados débilmente equivalentes a los TAGs \cite{vijay-shanker94}.
Por ejemplo, las Gramáticas Categoriales Combinatorias (CCGs) \cite{steedman00}, 
las Gramáticas Indexadas Lineales (LIGs) \cite{gazdar88} y las
\textit{Head Grammars (HGs)} \cite{pollard84}.
 
Trabajos posteriores dieron argumentos en contra de la adecuación de los TAGs.
Por ejemplo, para el idioma holandés, Manaster-Ramer \cite{manaster-ramer87}
propuso una construcción sintáctica que usa interdependencias seriales y
conjunciones.
Las conjunciones permiten agregar a la estructura básica de interdependencias
seriales $X^n Y^n$, nuevas series de verbos para obtener oraciones de la forma
\begin{quote}
... $X^n$ $Y^n$, $U^n$, $V^n$ \textit{en} (y) $W^n$.
\end{quote}
Un lenguaje con estas características no puede ser expresado por un TAG.

Sí existen otros formalismos MCS que logran expresar estas quíntuples
interdependencias, como por ejemplo los \textit{Linear Context-Free Rewriting
Systems (LCFRSs)} \cite{vijay-shanker87} y sus equivalentes \textit{Multiple
Context-Free Grammars (MCFGs)} \cite{kasami89}.
Estos formalismos son considerados los más representativos de los lenguajes MCS,
ya que no se han identificado hasta el momento otros formalismos \textit{mildly
context-sensitive} de mayor expresividad \cite{kallmeyer10}.

Una versión extendida del argumento de Manaster-Ramer con interdependencias
seriales propuso la inadecuación débil de los LCFRS \cite{groenink97}.
También se dieron argumentos contra la adecuación de los LCFRS tomados de los
idiomas alemán (mezclado de larga distancia) \cite{becker92}, chino (nombres de
números) \cite{radzinski91} y georgiano antiguo (apilado de casos)
\cite{michaelis97}, no sin algo de polémica.
En algunos de estos casos se afirma incluso que los argumentos de inadecuación
se aplican a todos los lenguajes MCS en general.

En vista de la posible inadecuación de los lenguajes MCS, también se propusieron
formalismos menos restrictivos, aunque siempre manteniendo la condición de
\textit{parsing} polinomial.
Entre ellos, se encuentran los \textit{Parallel MCFGs (PMCFGs)} \cite{kasami89},
los \textit{Simple Literal Movement Gammars (Simple LMGs)} \cite{groenink97} y
sus equivalentes \textit{Range Concatenation Grammars (RCGs)} \cite{boullier98}.
Los Simple LMGs y los RCGs generan exactamente la clase completa de lenguajes
con \textit{parsing} polinomial, denominada PTIME, por lo que son considerados
los formalismos más expresivos posibles para ser considerados adecuados para el
lenguaje natural.

\begin{figure}[tp]
 \centering
\begin{tikzpicture}[font=\scriptsize]
  \draw (-4, 0) -- (4, -0) -- (4, 4.5) -- (-4, 4.5) -- (-4, 0);
  \draw (1.5, 0) arc (0:180:1.5 and 0.5);  
  \draw (2.5, 0) arc (0:180:2.5 and 1.25);  
  \draw (3.5, 0) arc (0:180:3.5 and 1.75);  
  \draw (4, 0) arc (0:180:4 and 2.25) [style=dotted];  
  \draw (4, 1) arc (0:180:4 and 1.75);  

  \draw[pattern=dots] (4, 1.5) arc (0:90:8 and 2.5) -- (-4, 1.5)  
                  (-4, 1.5) arc (0:90:-8 and 2.5) -- (4, 1.5);  

  \draw (0, 0.25) node {CFG}; 
  \draw (0, 0.75) node {TAG = CCG = LIG = HG};
  \draw (0, 1.5) node {LCFRS = MCFG}; 
  \draw (0, 2) node {MCS};
  \draw (0, 2.5) node {PMCFG};
  \draw (-4, 3.5) node[anchor=west,align=left,fill=white,draw] {Simple LMG\\= RCG = PTIME};
  \draw (4, 3.5) node[anchor=east,align=center,fill=white,draw] {CSL};
\end{tikzpicture}
 \caption{
Detalle de la jerarquía de Chomksy entre los CFLs y los CSLs.
Se muestran algunos formalismos gramaticales discutidos en la literatura, y la
relación entre ellos en cuanto a poder expresivo.
La frontera moderadamente \textit{context-sensitive} (MCS) es difusa ya que no
existe una definición precisa.
No se conoce exactamente la relación entre PTIME y CSL, por lo que las áreas
punteadas no se sabe si son vacías o no.
}
 \label{fig:mildly}
\end{figure}
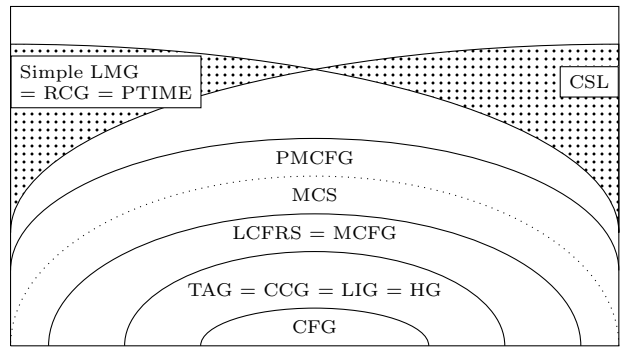

La relación entre los diferentes formalismos mencionados en esta sección puede
verse en el detalle de la jerarquía de Chomsky de la Fig.~\ref{fig:mildly}.
Obsérvese que no se conoce exactamente la relación entre PTIME y CSL.


\section{Discusión}
\label{sec:discussion}

La discusión de la relación entre el lenguaje natural y los lenguajes formales
permitió el desarrollo de una base teórica a donde se propusieron y estudiaron
numerosos formalismos gramaticales, de los que sólo mencionamos los más
relevantes en las secciones anteriores.

Esta discusión fue convergiendo hacia mediados de los 90, luego de haberse
llegado a cierto consenso acerca de la adecuación de los formalismos
moderadamente sensibles al contexto.
En esa época también hubo un fuerte viaje hacia los métodos empíricos basados en
datos, que supuso una pérdida de protagonismo de temas más teóricos como los
tratados en este artículo.

Por supuesto, muchos modelos empíricos desarrollados posteriormente se
beneficiaron de esta base teórica.
En particular, versiones probabilísticas de varios formalismos MCS fueron
utilizados para la implementación de parsers o traductores automáticos basados
en sintaxis.

Sin embargo, las principales líneas de investigación empírica siguen basándose
en formalismos que no superan el poder expresivo de las gramáticas libres de
contexto.
No existe un incentivo especial para intentar capturar fenónemos sensibles al
contexto, ya que los métodos de evaluación estrictamente cuantitativos
instaurados no los valoran especialmente.
Esto se debe principalmente a que los fenónenos sensibles al contexto son muy
poco frecuentes en los corpus de datos utilizados, e incluso en algunos casos
ni siquiera se encuentran anotados.

Es de esperar en algún momento el estado del arte alcance un punto en el que
la evaluación cuantitativa no sea lo suficientemente informativa.
Cuando esto suceda, se deberán incorporar criterios más cualitativos que
permitan apreciar el nivel de complejidad de los fenómenos sintácticos que los
sistemas son capaces de capturar.
De esta manera, es posible que la discusión sobre formalismos gramaticales más
adecuados vuelva a tomar un impulso e incluso empiece a ocupar un lugar
importante en el desarrollo de aplicaciones basadas en tecnologías de
Procesamiento de Lenguaje Natural.

\section*{Agradecimientos}

Agradezco a Miguel Pagano por la revisión detallada.
Sus comentarios fueron de gran ayuda para mejorar muchos aspectos del
artículo.
Este trabajo fue realizado con el apoyo de proyectos de la ANPCyT, Ministerio de
Ciencia, Tecnología e Innovación Productiva de Argentina (PICT 2014-1651)
y de la SECyT, Universidad Nacional de Córdoba.

\bibliography{main}

\end{document}